\documentclass[sn-mathphys-num]{sn-jnl}% Math and Physical Sciences Numbered Reference Style 
\usepackage{amsmath,amsfonts}
\usepackage{amsmath,amssymb,amsfonts}%
\usepackage{amsthm}%
\usepackage{mathrsfs}%
\usepackage{manyfoot}%
\usepackage{booktabs}%
\usepackage{algorithm}
\usepackage{algorithmicx}%
\usepackage{tikz}
\usepackage{algpseudocode}%
\usepackage{array}
\usepackage{csquotes}
\usepackage{listings}%
\usepackage{tabularx}
\usepackage{longtable}
\usepackage{mwe}
\usepackage{graphicx}
\usepackage{subcaption}
\usepackage{textcomp}
\usepackage{stfloats}
\usepackage{url}
\usepackage{verbatim}
\usepackage{graphicx}
\usepackage{subcaption}
\usepackage{multirow}
\usepackage{rotating}
\usepackage{caption}
\usepackage{array}
\usepackage{ragged2e}
\usepackage{hyperref}

\theoremstyle{thmstyleone}%
\theoremstyle{thmstyletwo}%

\theoremstyle{thmstylethree}%

\begin{document}

\title{HMR-ODTA: Online Diverse Task Allocation for a
Team of Heterogeneous Mobile Robots}
\author[1]{\fnm{Ashish} \sur{Verma}}\email{p20200020@pilani.bits-pilani.ac.in}

\author[1]{\fnm{Avinash} \sur{Gautam}}\email{avinash@pilani.bits-pilani.ac.in}

\author[1]{\fnm{Tanishq} \sur{Duhan}}\email{f20190636@pilani.bits-pilani.ac.in}

\author[1]{\fnm{V. S.} \sur{Shekhawat}}\email{vsshekhawat@pilani.bits-pilani.ac.in}

\author[1]{\fnm{Sudeept} \sur{Mohan}}\email{sudeeptm@pilani.bits-pilani.ac.in}

\affil[1]{\orgdiv{Department} \orgname{of Computer Science and Information Systems}, \orgaddress{\street{Birla Institute of Technology and Science}, \city{Pilani}, \postcode{333031}, \state{Rajasthan}, \country{India}}}

% \affil[2]{\orgdiv{School of Computing}, \orgname{University of North Florida}, \city{Jacksonville}, \postcode{32224},
% \state{Florida}, \country{USA}}

\abstract{Coordinating time-sensitive deliveries in environments like hospitals poses a complex challenge, particularly when managing multiple online pickup and delivery requests within strict time windows using a team of heterogeneous robots. Traditional approaches fail to address dynamic rescheduling or diverse service requirements, typically restricting robots to single-task types. This paper tackles the Multi-Pickup and Delivery Problem with Time Windows (MPDPTW), where autonomous mobile robots are capable of handling varied service requests. The objective is to minimize late delivery penalties while maximizing task completion rates. To achieve this, we propose a novel framework leveraging a heterogeneous robot team and an efficient dynamic scheduling algorithm that supports dynamic task rescheduling. Users submit requests with specific time constraints, and our decentralized algorithm—Heterogeneous Mobile Robots Online Diverse Task Allocation (HMR-ODTA)—optimizes task assignments to ensure timely service while addressing delays or task rejections. Extensive simulations validate the algorithm’s effectiveness. For smaller task sets (40–160 tasks), penalties were reduced by nearly 63\%, while for larger sets (160–280 tasks), penalties decreased by approximately 50\%. These results highlight the algorithm’s effectiveness in improving task scheduling and coordination in multi-robot systems, offering a robust solution for enhancing delivery performance in structured, time-critical environments.}

% \end{abstract}


\maketitle

\section{Introduction} \label{Introduction}
Multi-robot systems (MRS) have gained popularity due to their ability to perform complex tasks more efficiently than a single robot \cite{khamis2015multi}. MRS involves task distribution and parallel work toward a shared objective, using communication for coordination, which enhances efficiency and reduces task completion time \cite{vail2003multi}. However, MRS coordination presents challenges due to multiple non-trivial sub-problems such as task decomposition, multi-robot task allocation (MRTA), monitoring task execution progress, and termination detection. MRTA, which is NP-Hard \cite{gerkey2004formal}, focuses on collaborative task execution, such as in crucial areas like warehouse automation \cite{sharma2021coordination}, terrain coverage \cite{gautam2017fast, gautam2021multi}, exploration \cite{soni2022multi}, and search and rescue operations \cite{khairnar2023comparison}.

MRTA problem is one of the most challenging issues in Multi-Robot Systems (MRS), particularly when dealing with heterogeneous robots equipped with diverse sensors and actuators. These robots must optimally execute various tasks, each with distinct requirements and constraints \cite{Msala}. This problem can be viewed as an optimal assignment problem, where the goal is to allocate a set of tasks to a group of robots in a way that maximizes the overall performance of the system.

For the environments such as warehouses, hospitals, and office buildings often require the delivery of numerous items, presenting complex challenges that are commonly framed as Pickup and Delivery Problem (PDP). In this paper, we consider a real-world environment like a hospital building. In a hospital building, a team of heterogeneous robots is deployed to fulfill online incoming service requests from various users, such as doctors, wardens, and patients. Service requests are submitted via handheld devices or a web interface, with specified time constraints, and are stored in a global queue. Service requests include supply delivery, telemedicine, monitoring, patient assistance, transportation, logistics, and cleaning. In contrast to the Temporal Logic Task Allocation in heterogeneous multi-robot systems’ method addressed in \cite{Zavlanos}, which relies on centralized control and pre-specified task schedules, \textit{the proposed approach in this paper i.e., HMR-ODTA dynamically reallocates tasks through an auction-based mechanism. This ensures that even when tasks arrive unpredictably, they can be effectively handled without violating time constraints}. 

\textit{\textbf{Limitations of the State-of-the-Art:}} A recent review paper \cite{chakraa2023optimization}  highlights critical limitations in the field of dynamic task allocation, particularly in online and iterative methods for handling newly arriving tasks. These limitations stem from the inherent complexity of defining a cost function that can operate effectively in real-time under dynamic conditions. The challenge intensifies when task execution must adhere to strict temporal constraints, such as deadlines or specific time slots.

Addressing this gap, this paper introduces a novel approach viz., Online Diverse Task Allocation for a Team of Heterogeneous Autonomous Mobile Robots (HMR-ODTA). Using HMR-ODTA a team of heterogeneous autonomous mobile robots is deployed to respond to service requests that arrive iteratively and unpredictably. Each service request is associated with a deadline, with some tasks becoming infeasible post-deadline and others incurring penalties if delayed. The proposed approach enables efficient task allocation and execution in real-time, accommodating both hard and soft temporal constraints.

\textit{\textbf{Major contributions of the paper:}}
The major contributions of the paper are as follows:

\begin{itemize}

\item The proposed approach utilizes a more adaptive strategy by employing a Simple Temporal Network (STN) \cite{dechter1991temporal} framework to evaluate the availability of robots within the specified time interval, defined as the period between the service request's arrival time and its deadline. This method effectively addresses temporal constraints while incorporating essential factors—such as remaining energy, current load-carrying capacity, speed, and robot efficiency based on the type of service request—to optimize system performance.

\item HMR-ODTA also incorporated an adaptive task rescheduling mechanism within a robot’s schedule to handle delays or unforeseen tasks, ensuring minimal disruption and timely completion of service requests, which distinguishes it from recent approaches \cite{wang2023efficient, martin2021multi}. Moreover, it accounts for the types and number of service requests that can be assigned to each robot and dynamically adjusts task allocations to maximize the efficiency and utilization of heterogeneous robot teams.

\item The main objective of HMR-ODTA is to maximize the number of successfully completed service requests while minimizing associated penalties. By incorporating these considerations into the STN framework, the system efficiently balances trade-offs among energy consumption, the variety of service requests each robot can handle, and the need for dynamic rescheduling.

\item Extensive simulations using the ROS-based Gazebo simulator \cite{simulator} have been conducted to establish the efficacy of the proposed approach. Finally, two SOTA approaches, i.e., (a) Effective and Efficient Performance Impact \cite{wang2023efficient} (b) Multi-robot task allocation using Genetic Algorithm (GA-MR) \cite{martin2021event} are re-implemented for heterogeneous autonomous mobile robots for online PDP to be compared with the proposed HMR-ODTA algorithm. The proposed approach outperforms these SOTA approaches.
\end{itemize}

The following sections of this paper are organized as follows: Section \ref{section 2} provides an overview of the related work about the task assignment problem. Section \ref{preliminaries} describes the preliminaries of the proposed approach and symbols. Section \ref{section 3} delves into the formulation of the heterogeneous multi-robot service request assignment and execution problem. Section \ref{section 4} outlines the proposed decentralized auction algorithm. Section \ref{section 5} provides insights into the configuration of the simulation environment. Section \ref{section 6} presents the results followed by the conclusions in Section \ref{section 7}. 

\section{Related Work} \label{section 2}
Over the past decades, task delivery to customers has faced significant challenges due to the rise of e-commerce \cite{morganti2014impact}. Some environments, like warehouses, hospitals, and office buildings, require the delivery of numerous items. These challenges can be summarized as pickup and delivery problems (PDP), a type of vehicle routing problem (VRP) where items need to be transported from a pickup location to a drop-off location. PDP can be categorized into various types, such as synchronous PDP, asynchronous PDP, simultaneous multiple pickups and single deliveries, and simultaneous multiple pickups and multiple deliveries. An extension of PDP, known as multi-agent PDP (MAPDP), has been explored in recent works alongside the use of mobile robots \cite{wang2023efficient, Xiaoshan, martin2021event, wang2021task, Gong,Camisa}. MAPDP is an NP-hard problem \cite{hochba1997approximation}.

The problem of determining the optimal multi-robot coalition formation is well-known to be NP-hard \cite{khamis2015multi}. Existing literature on Multi-Robot Task Allocation (MRTA) emphasizes the use of teams of autonomous heterogeneous robots equipped with various actuators, sensors, and other components, making them capable of handling a wide variety of tasks \cite{Prorok}. Enhancing the heterogeneity of robot teams is therefore critical for achieving high-quality sub-optimal solutions \cite{Ferreira, wang2023efficient}. To address this challenge, \cite{Xiaoshan} introduces the Capacity-Constrained Heterogeneous Delivery Problem (CCHDP), which involves coordinating a single drone and a single truck. In \cite{Ferreira}, a decentralized auction-based algorithm is proposed to solve the Multi-Depot Vehicle Routing Problem (MDVRP), where robots are required to visit multiple nodes, complete tasks, and return to their depots. In \cite{wang2023efficient}, the Effective and Efficient Performance Impact (EEPI) algorithm is proposed, tackling a scenario where both robots and tasks are heterogeneous. However, each robot is restricted to executing a single type of task—for instance, some robots can only deliver food, while others can only deliver medicine. Additionally, the problem assumes that all tasks are predefined and governed by hard temporal constraints, meaning tasks that miss their deadlines are rejected outright.

In \cite{Dai}, a heuristic algorithm has been proposed to pick up fruits and come back to the depot with the help of homogeneous multi-robots, by using multi-picking robot task allocation (MPRTA). Unlike \cite{Dai}, in \cite{wang2021task}, two teams of specialized robots i.e., transport robots and pick robots work together to execute the multi-station order fulfillment tasks in a logistic environment, where the tasks are predefined to the heterogeneous robots. A heuristic algorithm based on game theory is proposed in \cite{Xiaoshan}, where the robots are assigned to deliver the packages to a group of customers. The robots are capable of carrying multiple packages simultaneously and delivering them sequentially. A novel method is proposed to provide a solution for MAPDP, where robots and humans work together in a shared environment \cite{Gong}. It is a two-phase algorithm, where the first phase is the task allocation, and the second phase is to calculate routes. An event mixed Integer linear programming (MILP) based algorithm is proposed to provide a solution for solar irradiance, where the robots will go to the locations and perform the required tasks such as dust cleaning, fixing the solar panels, etc., \cite{martin2021event}. In all the aforementioned algorithms \cite{Xiaoshan,martin2021event,wang2021task, Gong}, centralized organization-based algorithms have been proposed. However, the primary drawback of a centralized organization is the risk of a single point of failure \cite{Camisa}. Moreover, a static solution following PDP is impractical because it requires a rescheduling of all the tasks for the arrival of any new task \cite{Camisa}. Unlike the centralized approaches \cite{Xiaoshan,martin2021event,wang2021task, Gong}, a decentralized algorithm based on the primal decomposition method is proposed to solve the PDP \cite{Camisa}. In this approach, robots visit the pickup location to carry the payload and deliver it to the drop-off location, where the robots are homogeneous, and their allocated tasks are predefined. In contrast, a novel decentralized algorithm based on game theory for heterogeneous robots is proposed to visit the inspection locations and share critical information in hazardous environments \cite{Huo}. It is evident from the literature, there has been minimal research on addressing the multi-agent PDP (MAPDP), a pressing issue in real intelligent storage systems (ISS) \cite{dragomir2018multidepot}. The need to switch transportation modes in the multi-modality feature of realistic transportation networks was emphasized \cite{dragomir2018multidepot}. Existing MAPDP involve homogeneous tasks \cite{Xiaoshan, wang2021task,  Gong, Camisa}, with limited research addressing scenarios where both robots and tasks are heterogeneous for MAPDP\cite{wang2023efficient}. Moreover, in \cite{wang2023efficient}, tasks are already known beforehand. Therefore, it is essential to focus more on the development of online MAPDP algorithms that accommodate both heterogeneous robots and tasks.

In the literature, excessive research has been dedicated to addressing the MRTA problem \cite{khamis2015multi} - \cite{wang2021task}. MRTA can be classified into eight categories i.e., ST-SR-TA, ST-SR-IA, ST-MR-TA, ST-MR-IA, MT-SR-TA, MT-SR-IA, MT-MR-TA, MT-MR-IA \cite{gerkey2004formal}. Single Task Robots (ST) means each robot can perform at most one task at a time \cite{sakamoto2020routing}, whereas Single Robot Tasks (SR) states each task requires at most one robot \cite{sullivan2019sequential}. Multi-task robots (MT) are a sub-set of robots that can perform multiple tasks \cite{oliveira2021multi}, whereas Multi-robot tasks (MR) are a sub-set of tasks that require multiple robots. Time Extended Assignment (TA) allows task allocation to the robots based on already available information, i.e., available tasks and robots, and prediction on how tasks will arrive over time \cite{ferreira2021distributed}. Instantaneous Assignment (IA) allows task allocations based on available information concerning robots, tasks, and the environment, the solution allows instantaneous allocation of tasks to robots without considering the future allocations \cite{nanjanath2010repeated}. Based on the taxonomy proposed by \cite{gerkey2004formal}, ST-MR is the most preferable in the MRTA domain \cite{wang2021task}. 

According to the MRTA taxonomy \cite{gerkey2004formal}, the HMR-ODTA (Online Diverse Task Allocation for a Team of Heterogeneous Autonomous Mobile Robots) problem falls under ST-MR-TA. In offline scenarios, optimal solutions to Pickup and Delivery Problems (PDPs) are often achieved using branch-and-bound methods or heuristic approaches such as Tabu search \cite{parragh2006survey}, simulated annealing \cite{yu2016solving}, and genetic algorithms \cite{parragh2006survey}. PDPs typically include time constraints, requiring requests to be fulfilled within specified time windows. In online PDPs, however, requests arrive dynamically and are not known in advance. While static methods can be adapted to re-generate schedules as new information becomes available, limited attention has been given to managing dynamic events such as cancellations and delays \cite{XIANG2008534}. A review of the literature reveals a lack of research on decentralized cooperation for the Multi-Agent Pickup and Delivery Problem (MAPDP), particularly when involving teams of heterogeneous robots. To address these gaps, the authors propose an algorithm designed for heterogeneous robots capable of performing diverse tasks, such as delivering supplies or collecting data. Unlike prior works \cite{wang2023efficient, wang2021task, martin2021multi, Gong, Camisa, Dai}, the focus of the proposed algorithm is on online task allocation. To evaluate the algorithm, a hospital environment simulation model is developed using the Robotic Operating System (ROS), incorporating various characteristics of both robots and service requests. The algorithm's performance is validated through comparisons with state-of-the-art approaches, including Effective and Efficient Performance Impact (EEPI) \cite{wang2023efficient} and Multi-robot Task Allocation using Genetic Algorithm (GA-MR) \cite{martin2021multi}.

\section{Preliminaries} \label{preliminaries}
The proposed work investigates the deployment of a heterogeneous team of mobile robots to address a range of service requests within a hospital building while adhering to specified temporal constraints. These requests emanate from various entities within the hospital, like doctors, nurses, patients, caregivers, hospital pharmacists, assistants, etc. The requests relate to tasks such as supply delivery, telemedicine, monitoring, data collection, patient assistance, and more. These requests are stored in a global queue ($Q$) based on their arrival times. Upon arrival of a new service request, the robots are notified. Importantly, unlike the conventional method  \cite{quinton2023market}, the notifier does not directly assign service requests to the robots in the proposed approach.

The robots in the system adopt a round-robin approach (Algorithm \ref{Algorithm2}) to decide amongst themselves who will be responsible for conducting the auction for a given round. Once the auction is conducted, the request is added to the service request list (SRL) of the robot that wins the bid. The details about the robots, service requests, and auctioneer determination are given below:

\begin{itemize}
    \item \textbf{Service Request ID and Robot Classes:} Each service request is represented as a whole number $j$. The set of service requests is denoted as $J$. The system consists of a defined set of robot classes, denoted as CL = \{$CL_C$ and 0 $\leq$ C$ \leq$ m-1\}. Where m is the natural number representing the total number of robot classes and $CL_C$ represents the $C^{th}$ class. Further, each class consists of a number of robots. A robot belonging to class $C$ is identified as $R^r_C$ where $r$ and $C$ is a whole number.

    \item \textbf{Class Selection:}
    To determine the identity of the robot which will auction the $j^{th}$ service request, the following method is used: (i) the service request ID, denoted as $j$, undergoes a modulo operation with the number of robot classes, represented as $m$, (ii) this modulo operation produces a remainder value which determines the class of robots eligible to participate in the auction for that specific service request $j$.

    % \newpage%
    \item \textbf{Robot Selection within Class:} Once the class of robot for auctioning a service request is determined, the service request ID, $j$, is further divided by the number of robot classes. This is an integer division and a quotient is obtained. Next, a modulo operation is again performed between the obtained quotient and the number of robots within the selected class. This second modulo operation produces another remainder value. This final remainder value specifies the particular robot from the selected robot class that will be responsible for conducting the auction for the service request $j$.

\end{itemize}

Each robot maintains an auction queue (${aucq}{R^{r}_{C}}$) to store service requests eligible for bidding. The auctioneer robot broadcasts the new service request to other robots. Upon receiving the notification from the auctioneer, all robots, including the auctioneer itself, compute their bid values as outlined in Algorithm \ref{Algorithm3}.

The details about the service request scheduling for an individual robot are presented below:
% \newpage

\begin{itemize}
   
\item \textbf{Schedule Calculation:} The robot calculates a schedule using the Simple Temporal Network (STN) \cite{dechter1991temporal} for all the service requests in its list. The STN generates a schedule based on factors like service request attributes, robot capabilities, and constraints (refer to Section \ref{STN}).

\item \textbf{Executing Service Requests:} The robot starts executing service requests as per the schedule generated by the STN. It starts at the first node (i.e., task) of the schedule.

\item \textbf{Updating Current State:} After executing a service request in the schedule, the robot updates its current state, which includes its location and current capabilities, such as its load carrying capacity, remaining energy, etc.

\item \textbf{Checking for New Service Requests:} The robot continuously monitors for new service requests. By checking its Service Request List (SRL).

\item \textbf{Schedule Recalculation:} If there are updates in the Service Request List (SRL) at time instant $t+k$ (where $t,k$ is the start time and time needed to perform a task, respectively), the robot dynamically re-schedules its task execution list using STN. This allows the robot to adjust its existing schedule to accommodate any new service requests. If there are no updates to the SRL, the robot continues to follow the previous schedule until all service requests in the schedule are visited.

  \item \textbf{Completion and Waiting:} The robot continuously checks the SRL for any new updates. If there are no updates, the robot waits for new service requests to arrive.
  
\end{itemize}

The bid value calculation of an individual robot depends on its schedule. If the robot cannot add that service request to its SRL due to capacity constraints or other reasons (e.g., temporal constraints), then it will submit a bid value with an exceptionally high value (= $\infty$) to indicate its inability to fulfill the request. The service requests will be allocated to the robots with the best bid. The bid calculation method is discussed in Section \ref{BestBid_Algorithm}.

\section{Problem Definition} \label{section 3}

We consider a fleet of heterogeneous robots belonging to a set of m different classes $CL = \{CL_0, CL_1, $...$, CL_{m-1}\}$. These robots are employed to perform different types of service requests. The different types of service requests are denoted by the set by $T = \{T_{1}, T_{2}, $...$, T_{P}\}$, where $P$ is the cardinality of different types of possible service requests. There are $x$ different depots that can cater to robots belonging to any class. The set of depots is represented as $D = \{D_1, D_2,$...$, D_{x}\}$. Initially, all robots are assumed to be at a docking station. Robots must deliver a set of service requests, $ J$, that are not known beforehand. Each service request has distinct requirements for its completion. To fulfill a service request, a robot needs to first reach the pickup location and subsequently move on to the drop-off location. 

A service request is considered to be completed if the robot servicing it reaches the designated drop-off location. The key objectives of allocating service requests are (i) to determine which robot will carry out what service request and (ii) what will be the order in which a robot will attend to the service requests allocated to it. The objective of this allocation is to improve global outcomes, with a specific focus on a critical metric, i.e., minimizing the overall penalties associated with carrying out these service requests. For the task allocation problem addressed in this paper, it is assumed that each robot can handle only one service request at a time, and each service request requires a single robot for its execution. This problem falls under the classification of ST-MR-TA. The following are the assumptions made:

\begin{itemize}

% \item Each robot (agent) has its velocity $(v)$ and certain payload capacity $(C)$. 

\item The robots possess the capability to localize and are equipped with sensors that enable them to detect obstacles as well as other robots present in the environment.

\item Each service request is treated as an individual, and the same type of requests may have different attributes as described in section \ref{Service Requests}.

\item The robots operate without prior knowledge of the specific service requests that users might make. 

% However, it is assumed that incoming service requests will align with the parameters as described in Table  \ref{Table4}.

\item Robots in a class have the same attributes as described in the section \ref{Robot Model Team}.

\item Robots have the flexibility to recharge at any depot, and multiple robots can simultaneously recharge at a single depot. Furthermore, the charging duration remains uniform across all robot classes. The charging time for each robot is fixed, i.e., 300 sec.

\end{itemize}

Based on the above assumptions, we propose an MRTA algorithm called HMR-ODTA, designed for distributed online task (service request) allocation. 

\subsection{Environment Model} 
\begin{figure} [htbp]
    \centering
    \includegraphics[width=\textwidth]{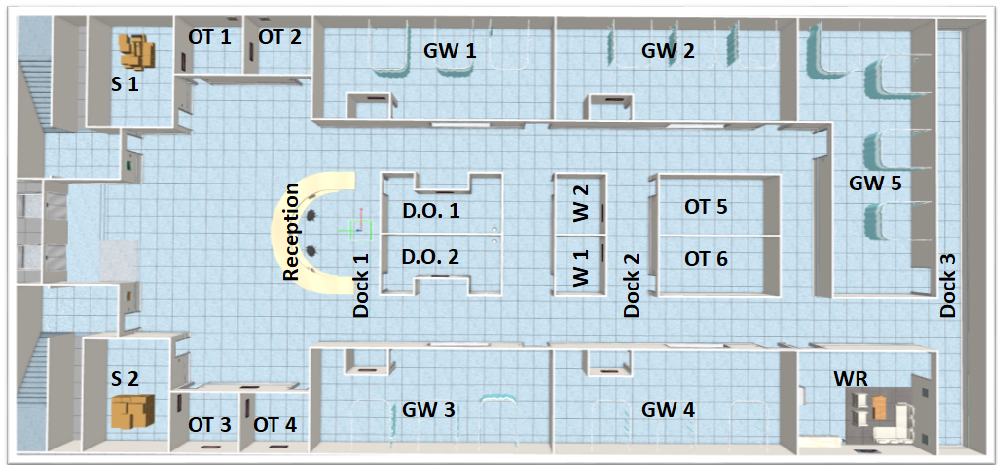}
     \caption {Top view of the hospital building \cite{Hospiotal}. Labels: OT-Operation Theatre, GW-General Ward, D.O.- Doctor Office, S-Store, W-Washroom, WR-Waiting Room.} \label{Hospital Building}
  \label{Fig: 2}
\end{figure}

We consider a fleet of robots operating in a hospital building. The workspace is mapped as an occupancy grid as shown in Fig. \ref{Hospital Building}. Mapping the environment helps the robot to navigate to specific locations, such as the reception area, emergency room services, operation theatre, etc. Occupancy grid map facilitates various aspects of mobile robot navigation, such as localization, path planning, and collision avoidance  \cite{thrun2001learning}.

\subsection{Service Requests Properties}  \label{Service Requests}
  The robot must deliver a set of service requests $J = \{j^1, j^2,..., j^i\}$ that are not known beforehand. Every service request might have distinct properties and be treated as an individual. Service requests can originate from any of the labels shown in Fig. \ref{Fig: 2}. Each service request is associated with a specific type $(T_{p})$. Each service request has eight attributes, i.e., $j^i = \{j_{PL},j_{DL}, j_{T_p},  j_{dem}, j_{AT}, j_{ET}, j_{DD}, j_{TC} \}$, where $j^i \in J$.

The details of the eight attributes are as follows:

\begin{itemize}
%\item $T_{ij}$: represents the $i^{th}$ number of $j^{th}$ type service request.
\item $j^i$: represents the id of the $i^{th}$ service request, where $i$ is a natural number.
\item $j_{PL}$: represents the pickup location.
\item $j_{DL}$: represents the drop-off location.
\item $j_{T_{p}}$: represents the $p^{th}$ type of service request, where $T_p \in T$.
 % for set $\{a1, a2, …, an\}$.
\item $j_{dem}$: represents the demand in $kg$.
\item $j_{AT}$: represents the arrival time of service request. 
\item $j_{ET}$: represents the earliest pickup time of service request.
\item $j_{DD}$: represents the delivery deadline in seconds.
\item $j_{TC}$: represents the type of temporal constraint (Hard / Soft). 

\end{itemize}

In this paper, we consider both hard and soft time constraints for service requests. If none of the robots can execute a service request within the specified time window and the type of temporal constraint is hard, then such service requests are rejected. On the other hand, for service requests with soft time constraints, if the service request violates the time window, then the service request is still accepted but with a penalty. The penalty function $j_{Pen}$ is defined in Eq. \ref{equation 1}:

\begin{equation}
     j_{Pen} = (\max \{0, {(j_{CT} - j_{DD})}\}) \label{equation 1}
\end{equation}

\noindent Where $j_{Pen}$ signifies the penalty associated with the delayed execution of the $j^{th}$ service request. $j_{CT}$ and $j_{DD}$ are the completion time and delivery deadline time of the $j^{th}$ service request, respectively. This implies that if a service request is executed within its designated deadline, there will be no penalty. Conversely, if a service request exceeds its deadline, the penalty will be increased with increasing task completion time.

\subsection {Robot Model and Team} \label{Robot Model Team}

Initially, all robots are assumed to be stationed at the depots. We examine various types of service requests which require a team of diverse robots to execute. Each robot r, belonging to class C has eight attributes represented as $R^{r}_{C} = (A_{R_C}, v_{R_C}, B_{R_C}, aucq_{R^r_C}, SRL_{R^r_C}, P_{R_C}(v), E_{R_C}, W_{R_C})$. The details of the eight attributes are as follows:

\begin{itemize}

\item $A_{R_C}$: the set of types of service request $\{T_{1}, T_{2}, $...$, T_{p}\}$ can assign to robots of $C^{th}$ class.
 % for set $\{a1, a2,$...$, an\}$.
\item $v_{R_C}$: the speed of robots of the $C^{th}$ class.
\item $B_{R_C}$: the maximum weight the robots of $C^{th}$ class can carry at any time.
\item $aucq_{R^r_C}$: the auction queue of the $r^{th}$ robot of $C^{th}$ class.

\item $SRL_{R^r_C}$: the service requests list (SRL) stores the id of the service requests won by the $r^{th}$ robot of $C^{th}$ class. 
\item $P_{R_C}(v_C)$: the power consumed by the robots of $C^{th}$ class traveling at velocity $v_C$.
\item $E_{R_C}$: initial energy of the robots of $C^{th}$ class.
\item $W_{R_C}$ is the weight of the robots of $C^{th}$ class.

\end{itemize}

\subsection{Energy Consumption}

Initially, the robots start with a full charge, having an initial energy level of $E$ joules. The estimated travel time $TT$ for a robot is determined by considering its initial energy and the power required to sustain its velocity while not carrying any payload. This is calculated as in Eq. \ref{equation 2}:

\begin{equation}
TT_{R_{C}^{r}} = \frac{E_{R_{C}}}{P_{R_{C}^{r}}(v_C)} \label{equation 2}    
\end{equation}
Here $TT_{R_{C}}$, $P_{R_{C}}(v_C)$ are the estimated travel time and power consumed by $r^{th}$ robot of $C^{th}$ class to maintain its velocity $v_C$ \cite{mei2004determining}. $E_{R_{C}}$ is the initial energy of the robot of $C^{th}$ class.

Robots carrying different payloads consume different energy to maintain the same velocity. Therefore, the energy consumption per unit time for different classes of robots will be different. Robots carrying different payloads consume different amounts of energy to maintain the same velocity. Therefore, energy consumption for different classes of robots will be different \cite{shuang}. The energy consumption $(E1)$ for which the robot navigates from the current location to the pickup location of any service request is formalized below in Eq. \ref{e1}:

\begin{equation}
% \Large
\begin{cases}
E1_{R_{C}^{r}} = (\frac{1}{2} * v_C^2 * W_{R_{C}}  + \mu * g * v_C \\ * W_{R_{C}} * t1) \\
\end{cases}
\label{e1}
\end{equation}

Where $W_{R_C}$ represents the $C$ th class robots' weight. $v_C$ is the velocity $(m/s)$ of the $C$ th robot class, $\mu$ is the friction coefficient ($\mu = 0.4$), $t1$ is the time taken to travel from the current location to the pickup location and $g$ is the acceleration due to gravity.

The energy consumption $(EC)$ for which the robot navigates with payload is formalized below in Eq. \ref{equation 3}:

\begin{equation}
\begin{cases}
EC_{R_{C}^{r}} = 0,  \text{if idle} \\
EC_{R_{C}^{r}} = (E1_{R_{C}^{r}}+\frac{1}{2} \times (v_{R_C})^2 + \mu \times g \times v_m \times t  ) \\
\quad \times \left(W_{R_{C}} + \sum _{j=1}^{\lvert SRL_{R_{C}^{r}} \rvert} j_{\text{dem}} \times x_{j}\right), \text{if moving}
\end{cases}
\label{equation 3}
\end{equation}

\noindent Where $E1_{R_{C}^{r}}$ is calculated from the Eq. \ref{e1}. $W_{R_C}$ represents the weight of the robots of $C^{th}$ class, $j_{dem}$ is the demand of the $j^{th}$ service request. $SRL_{R_{C}^{r}}$, $\lvert SRL_{R_{C}^{r}} \lvert $ represents the service request list and the number of allocated service requests to the robot of id $R_{C}^{r}$, respectively. $v_{R_C}$ is the velocity $(m/s)$ of the $C^{th}$ robot class, $\mu$ is the friction coefficient ($\mu = 0.02$), $t$ is the time taken to travel from the current location to the next node and $g$ is the acceleration due to gravity. $x_{j} \in \{0,1\}$.
\begin{itemize}
\item $x_{j}$ = 0; if the $j^{th}$ service request payload is not loaded.
\item $x_{j}$ = 1; if the $j^{th}$ service request payload is loaded.
\end{itemize}

\noindent Robots need to charge themselves periodically at the docking station. The minimum energy ($ME$) required to return to the nearest docking station is calculated as in Eq. \ref{equation 4}: 

\begin{equation}  \label{equation 4}
   ME_{R_{C}^{r}} = \frac{P_{R_{C}^{r}}(v)* \min (d(L_{R_{C}^{r}}, D_x)}{v_{R_C}} 
\end{equation}

\noindent Where $L_{R^{r}_{C}}$ is the current location of the robot $R^{r}_{C}$, $P_{R_{C}^{r}}(v)$ is the power to maintain velocity $v$, and $d(L_{R^{r}_{C}}, D_x))$ represents the geodesic distance from the robot's current position to the nearest docking position. The geodesic distance between all the depicted points in Fig. \ref{Hospital Building} are calculated using the improved jump point search path optimization algorithm (JPS) \cite{luo2022improved}.

\subsection{Robot Efficiency In Terms Of Types Of Service Request}
The efficiency of a robot in performing different types of service requests is calculated as in Eq. \ref{equation 5}:

\begin{equation}  
     \eta_{R_C} = \frac{\:\mid A_{R_C}\mid}{\lvert T \lvert} \label{equation 5}
\end{equation}

\noindent Where $\mid A_{R_C} \mid$ denotes the cardinality of the set $A_{R_C}$. $A_{R_C}$ denotes the set of types of service requests that can be assigned to a robot belonging to $C^{th}$ class. $ \eta_{R_C}$ represents the efficiency of the $C^{th}$ class robot ($R_C$). $\lvert T \lvert$ is the total number of types of service requests possible in the system. This calculation gives a ratio between the number of different types of service requests that the robot can perform and the total number of different types of service requests possible in the system. The resulting number can be expressed as a percentage to indicate the robot’s efficiency in performing a range of different types of service requests. It is important to note that the efficiency of the two classes of robots can be equal even though their abilities differ.

\subsection{Constraints} \label{section Validates Service Requests}
The different constraints arising from the problem description are as follows.
\begin{itemize}

   \item The type of incoming service request must map to at least one of the robot classes from the available $C$ robot classes.

    \item The demand or payload $(j_{dem})$ of the $j^{th}$ service request should be within the robot’s load carrying capacity $(B_{R_C})$. This constraint is defined in Eq. \ref{equation 6}:
    
    \begin{equation}  \label{equation 6}
   \begin{array}{l}
       \forall \: j \in N    : j_{dem} \leq B_{R_C}
    \end{array}
    \end{equation}

    \item The sum of payloads of service requests assigned to robot $R^{r}_{C}$ is required to be less than $B_{R_{C}}$ due to the robot's limited capacity. This constraint is defined in Eq. \ref{equation 7}:
     \begin{equation}  \label{equation 7}
         \sum_{j=0}^{\mid SRL_{R^{r}_{C}}\mid} {j_{dem} \leq B_{R_{C}}}
    \end{equation}
    
   $SRL_{R^{r}_{C}}$ represents the service request list (SRL) of the robot $R^{r}_{C}$.

   \item  Each service request can only be assigned to a single robot. It is defined in Eq. \ref{equation 8}:
\begin{equation} \label{equation 8}
 \begin{array}{l}
        SRL_{R^{r}_{C}} \cap SRL_{R^{r'}_{C'}} = \emptyset, \quad \forall \, (r,r') \in r^2, \\
        \, \forall \,(C,C') \in C^2
    \end{array}
\end{equation}
     such that if $(r = r')$ then $(C \neq C')$ or vice-versa.

\item Service request $j^i$ can be assigned to a robot $R^{r}_{C}$, as defined in Eq. \ref{equation 9}.
\begin{equation} \label{equation 9}
    j_{T_{p}} \in A_{R^r_C}
\end{equation}
$A_{R^r_{C}}$ represents the set of different types of service requests that can be assigned to $R^r_C$.
\end{itemize}

\subsection{Objectives}
The objectives of HMR-ODTA are as follows:
\begin{itemize}

\item Minimize the total penalty $(Pen)$ incurred for not completing tasks within the deadline. It is defined in Eq. \ref{equation 10}:

\begin{equation} \label{equation 10}
    min \sum_{m=0}^{m-1}\sum_{r=0}^{\lvert R_{m}\lvert -1}\sum_{j=1}^{\mid SRL_{R^r_{m}}\mid} ({j_{Pen}})* x_{SRL_{R_{m}^{r}}} 
\end{equation}

Where $m$ is the total number of robot classes, $\mid R_m \mid$ represents the total number of robots belonging to the $m^{th}$ class, and $SRL_{R_{m}^{r}}$ stores the id of the service requests won by the $r^{th}$ robot of $m^{th}$ class.$j_{Pen}$ represents the penalty for executing $j^{th}$ service request (refer to Eq.\ref{equation 1}).

\begin{itemize}
\item $x_{SRL_{R_{m}^{r}}}$ = 0; if $j$ $\notin$ $SRL_{R_{m}^{r}}$.
\item $x_{SRL_{R_{m}^{r}}}$ = 1; if $j$ $\in$ $SRL_{R_{m}^{r}}$.
\end{itemize}

\end{itemize}

\section{Proposed Approach} \label{section 4}

We propose a decentralized Multi-Robot Task Allocation (MRTA) algorithm that involves allocating service requests to robots in a way that optimizes the total incurred penalty while considering the capabilities of the team of heterogeneous robots and the time constraints of the service requests. 

\subsection{The Proposed Architecture} \label{Proposed Architecture}

We present the framework for HMR-ODTA, Heterogeneous Mobile Robots - Online Diverse Task Allocation method. This approach is specifically designed to improve coordination within a team of heterogeneous robots when they undertake the execution of service requests. In this approach, each robot is responsible for making its own decisions and communicating with other robots to coordinate their actions. At the same time, all the robots are notified about any new service request from the web server that accepts service requests from the users. By allowing each robot to make its own decisions and communicate with others, the HMR-ODTA reduces the need for centralized coordination. The use of a server to check the status of service requests further enhances coordination efficiency by allowing all robots to have a shared understanding of the system’s current state. This helps prevent conflicts and ensures the requests are completed efficiently and effectively. In the proposed architecture, every robot maintains its individual auction queue, which is used to store information related to service requests intended for auction. The components of the architecture are described below:
  
\begin{itemize}
  
   \item \textbf{Map Representation}:- We create a topological graph representation of the environment where service requests (nodes) and depots are located (hospital). Each point is a node in the graph, and the distances between them are the edges. We use the jump point search (JPS) path planning algorithm to find the costs of the edges based on the geodesic distance between the nodes \cite{hart1968formal}.
   
    \item \textbf{Web Server (Global Queue)}:- Service requests arriving from the users are stored in a global queue maintained by the web server. The purpose of the global queue ($Q$) is to reduce redundancy in processing the service request and improve the system's overall efficiency. Another advantage of using a global queue is the reduced overall memory requirement. Only one copy of the status data needs to be stored in the global memory, reducing memory requirements. This improves the overall system scalability.
    
    \item \textbf{Robot Parameters}:- A robot is characterized by information such as the robot's ID $(R^r_{C})$, its abilities, attributes (described in section \ref{Robot Model Team}), current location in reference coordinate system and status (e.g., idle, in use, charging). The status of the robots can be:
    \begin{itemize}
        \item $\textbf{Busy}$: A robot is \enquote{$\textbf{busy}$} when it has a request in its service request list (SRL) on which it is currently working or the robot is conducting an auction. This means that it is actively performing a service request and can schedule new requests until its carrying capacity is not full (refer to Eq. \ref{equation 7}).
        \item $\textbf{Free}$: A robot is said to be \enquote{$\textbf{free}$} when its SRL is empty and the robot is available to receive new service requests.
        \item $\textbf{Failed}$: A robot is considered \enquote{$\textbf{failed}$} when it is unable to perform an allotted service request due to hardware or software malfunction. In this case, the robot may require maintenance or repair before it can resume normal operation.
        
        \item $\textbf{In charging mode}$: A robot is considered to be \enquote{$\textbf{in charging mode}$} when it has reached its minimum energy level $(ME)$ (refer to Eq. \ref{equation 4}).
        
    \end{itemize}
 
    \item \textbf{Service Request Parameters}:- The service request parameters include information such as the service request's ID $(j)$, its requirements (e.g., pick up a package, deliver it to a specific location), its status (e.g., pending, in progress, completed, rejected), and a timer that tracks how long has the service request been waiting or how long has the request been in progress. The status of the service request can be:
    \begin{itemize} \label{Robot State}
    
        \item $\textbf{Pending}$: A service request is \enquote{$\textbf{pending}$} when it is in the global queue or robot's auction queue ($aucq$).
        
        \item $\textbf{In progress}$: A service request is \enquote{$\textbf{in progress}$} when assigned to a robot and is currently being worked on or in the robot's service request list (SRL).
        
        \item $\textbf{Completed}$: A service request is considered \enquote{$\textbf{completed}$} when the robot has completed it and has removed it from its SRL. 

        \item $\textbf{Rejected}$: A service request is considered \enquote{$\textbf{rejected}$} when it has a hard time constraint and cannot be completed by any robot within the specified deadline.
        
    \end{itemize}
    
    The robots maintain an auction queue consisting of service requests that need to be auctioned. Once the service requests are stored in the robot's auction queue, the robot proceeds to initiate an auction for these service requests (refer to Algorithm \ref{Algorithm2}). The robots engage in negotiations amongst themselves, ultimately assigning the service request to the most efficient robot, as detailed in Algorithm \ref{Algorithm1}. Subsequently, the service request is included in the winning robot's service request list. The robot's service request list consists of a set of service requests that the robot is obligated to fulfill. The robot can establish priorities among its service requests, ensuring the successful execution of each one of them. When the robot finishes the execution of a service request from its list, it removes the completed request and proceeds to the next one in sequential order until it exhausts the complete list. In the event, that the list becomes empty, the robot will maintain its current position, either awaiting new service requests for assignment or evaluating its energy threshold (refer to Eq. \ref{equation 4}).
 \end{itemize}

\subsection{Managing Service Request List (SRL): Simple Temporal Network} \label{STN}
To effectively manage the SRL, each robot generates a Simple Temporal Network (STN) \cite{dechter1991temporal}. A Simple Temporal Network is a form of a directed graph in which each node corresponds to an action, and each edge denotes the time difference between two nodes. If an edge exists from node X to node Y, it signifies that action X should be finished before action Y begins. The weight of the edge represents the minimum duration required between X's completion and Y's beginning. One crucial characteristic of an STN is that it should not have any cycles since it would indicate a conflict in the constraints. This inconsistency would render it impossible to determine a viable solution. We use the STN concept to represent the inter-relatedness between various service requests required to meet the demands while complying with specific limitations such as deadlines, and resource availability. The scheduling for the start and completion time of a hard-time-constrained service request ($j$) is defined as follows:

\begin{itemize}
      \item The start time should be scheduled within the interval [$j_{ET}, j_{DD} - \frac{d(L_{R_{C}^{n}}, j_{PL})}{v_C} - \frac{d(j_{PL}, D_x)}{v_C} - \textit{charging time} - \frac{d(D_x,j_{DL})}{v_C}$], where $\frac{d(j_{PL}, D_x)}{v_C}$ and $\frac{d(D_x,j_{DL})}{v_C}$ represents the time taken by robot $R_{C}^{r}$ to travel from the pick-up location ($j_{PL}$) to the nearest charging station (depot $D_x$) and charging station to drop-off location ($j_{DL}$) at velocity $v_C$, respectively. Also, $\frac{d(L_{R_{C}^{n}}, j_{PL})}{v_C}$ represents the time taken by the robot to travel from its current location to the pickup location.  $j_{ET}$, $j_{DD}$ is the earliest pick-up time and deadline time of $j^{th}$ service request.

    \item The completion time, denoted as $j_{CT}$, should be scheduled within the interval [$j_{ET} +\frac{d(L_{R_{m}^{n}}, j_{PL})}{v_C} + \frac{d(j_{PL}, D_x)}{v_C} + \textit{charging time} + \frac{d(D_x,j_{DL})}{v_C}, j_{DD}$].
    
\end{itemize}

\noindent The above time intervals ensure the completion of a service request ($j$) within its specified deadline ($j_{DD}$) while taking into account the energy limitations of the robot. In cases where the robot lacks sufficient energy or when the time required to travel between the pick-up and drop-off locations is less than the robot's estimated travel time (refer to Eq. \ref{equation 2}), an alternate route is employed. This alternate route involves going from the pick-up location to a charging station and then proceeding to the drop-off location. This scheduling strategy aims to guarantee that the service request is fulfilled within the specified deadline, while simultaneously considering the robot's energy constraints.

When a new service request enters the environment, all compatible robots examine every available position within the corresponding layer of the STN to determine the optimal placement for the service request. Subsequently, these robots submit bids to compete for the execution of the given service request. This process ensures that the service request is efficiently allocated to a suitable robot within the constraints of the STN. The Floyd-Warshall algorithm is used to solve the STN in $O(s^3)$ polynomial time, where $s$ is the number of time points in the network \cite{lin2003new}.

\subsection{Service Request (Re)Insertion (Algorithm \ref{Algorithm1})}

When a service request is received at any instant, robots fetch it from the global queue ($Q$) into their auction queue ($aucq_{R_{C}^{r}}$).  The round-robin algorithm determines the auctioneer (refer to Algorithm \ref{Algorithm2}). It is worth noting that the server is not the auctioneer. The server is just an interface between the users and the robots. The auctioneer adds the service request into its auction queue (Line \ref{append}). The auctioneer announces the service request (Line \ref{Broadcast}), and all the robots (including the auctioneer) calculate their respective bid value and send it to the auctioneer. The auctioneer then determines the best bid value (Line \ref{start bidcalculation} to Line \ref{end bidcalculation}) and assigns the service request to the robot having the best bid (refer to Algorithm \ref{Algorithm4}). Finally, the winning robot adds the service request to its service request list (line \ref{Add Request}). Also, before inserting service requests into the server, prior validation of service requests is required (described in Section \ref{section Validates Service Requests}).

\begin{algorithm}[h]
\caption{Service Request (Re)Insertion Algorithm}\label{Algorithm1}
\small % Adjust font size
\begin{algorithmic}[1]
\Ensure For all $j \in J$
\Require $j \gets \text{ValidatedServiceRequest}(j)$ (Section\ref{section Validates Service Requests})
\State $\text{Q.append}(j)$

\If{$\text{size}(Q) \neq \emptyset$}
    \If{$j$ is unassigned}
        \State $ R^{r}_{C} \gets \text{AuctioneerDetermination}(j)$  \label{AuctioneerDetermination}
        \State $ aucq_{R_{C}^{r}}.\text{append}(j)$  \label{append}
        \State $R^{r}_{C}.\text{Broadcast}(j)$ \label{Broadcast}
        \For{$i = 0, \ldots, {m-1}$} \label{start bidcalculation}
            \For{$r = 0, \ldots, \lvert R_{C} \rvert - 1$} \label{secondloop}  
                \State $ R^{r}_{i}, Bid \gets \text{ComputeBid}(R^{r}_{i}, j)$ \label{ComputeBid}
                \State $ (Bid', R^{r'}_{i'}) \gets\text{BestBid}(R^{r}_{i}, Bid)$ \label{BestBid}
            \EndFor
        \EndFor  \label{end bidcalculation}

        \If{$Bid' \neq \infty$}
            \State Assign $ j$ to the robot $ R^{r'}_{i'}$ \label{Assign Request}
            \State $ R^{r'}_{i'}.SRL_{R^{r'}_{i'}}.\text{add}(j)$ \label{Add Request}
            \State $\text{Q.remove}(j)$
        \Else
            \State Reject $j$ \label{Reject Request}
            \State $\text{Q.remove}(j)$
        \EndIf

        % \State \hspace{0.5cm} \hspace{0.5cm} Update the status of the robot $R'_{mn}$ \label{update}
    \EndIf
\Else
    \State \textbf{Wait for server notification} \label{waiting}
\EndIf
\end{algorithmic}
\end{algorithm}

\subsection{Auctioneer Determination (Algorithm \ref{Algorithm2})} 

An example of the auctioneer determination  using Algorithm \ref{Algorithm2} is as follows:

Let $j = 1$ and $m = 4$, then $1\mod 4 = 1$, then value $X = 1$ determines that the robot belongs to the $R_{1}$ class and will conduct the auction (Line \ref{Start RR}). Suppose there are \(20\) robots in $R_{1}$ class ($\lvert R_1 \lvert$ = \(20\)). Therefore, the value $Z = 0$, which means the first robot of class $R_{1}$ (i.e., $R^{0}_{1}$) will perform the auction (refer to Line \ref{quotient} - Line \ref{end RR}).

\begin{algorithm}[h]
\caption{Auctioneer Determination ($j$)}\label{Algorithm2}
\begin{algorithmic}[1]
\State j $\gets$ $j^{th}$ service request \;
\State m $\gets$ $\lvert R \rvert $; \: \: \textit{/* Number of robot classes */} 
\State $ R_{X} \gets j \mod m $ \label{Start RR}
\State $ Z  \gets  \lfloor {\frac{j}{m}} \rfloor $\;  \label{quotient}
\State $ R^Y_{X} \gets Z \mod  \lvert R_{X} \rvert $\;  \label{end RR}
\State \textbf{return} $  R^Y_{X} $
\end{algorithmic}
\end{algorithm}

\subsection{Bid Calculation Process (Algorithm \ref{Algorithm3})} \label{Bid calculation}

In this section, we present how the bid value is returned from the STN (discussed in Section \ref{STN}). The STN takes into account the attributes of both the service request and the robot. It then evaluates various scenarios or combinations for incorporating the new service request into the robot's existing schedule. These scenarios may involve different schedules or sequences in which the robot can handle multiple service requests. For each of the evaluated scenarios, the STN calculates a bid value. This bid value is a composite measure that likely includes several components: the overall penalty, completion time ($CT$), efficiency ($\eta$), and remaining energy. Among all the calculated bid values for the different scenarios, the STN identifies and selects the \enquote{best local bid value}. This best bid value represents the most favorable or advantageous scenario for inserting the new service request into the robot's schedule within the immediate local context. It may not necessarily be the globally optimal solution, but rather the best option considering the attributes and constraints at hand. It is important to note that while the \enquote{best local bid value} may not guarantee a globally optimal solution; it provides a practical and context-aware choice for scheduling the service request based on the given criteria and available resources for that specific robot. 

\begin{algorithm}[h]
\caption{ComputeBid($R^{r'}_{C}, j$)} \label{Algorithm3}
\small % Adjust font size
\begin{algorithmic}[1]
\State $\textit{penalty} \gets \infty$
% \State $\textit{tt} \gets \infty$
\State $\textit{CT} \gets \infty$
\State $ \eta \gets \eta_{R_{C}}$
\State $\textit{UsedEnergy} \gets 0$
\State $\textit{temp} \gets \textit{SRL}_{R^{r'}_{C}}.\textit{add}(j)$
\State $\textit{currPos} \gets R^{r'}_{C}.currPos$
\State $\textit{Bid}(\textit{penalty}, \textit{CT}, \: \eta, \textit{UsedEnergy}, \textit{energyRem}) \gets \textit{STN}(\textit{currPos}, \textit{temp})$
\State \textbf{return} $\textit{$R^{r'}_{C}$, Bid}$ \label{bestRobot}
\end{algorithmic}
\end{algorithm}

\subsection{Finding the Best Bid Process (Algorithm \ref{Algorithm4})} \label{BestBid_Algorithm}

This section discusses the various criteria for finding the best bid. This algorithm defines a function called BestBid, which takes a bid as input and aims to determine the best bid and the corresponding robot. There is a conditional check that compares the current bid ($Bid$) to the best bid found so far ($BestBid$) based on various criteria:

\begin{algorithm}[h]
\caption{BestBid($R^{r'}_{C}$, Bid)} \label{Algorithm4}
\begin{algorithmic}[1]
\State $\textit{BestBid} \gets \textit{prev\_bestBid} $
\State $\textit{BestRobot} \gets \textit{prev\_bestrobot}$
\State $\textit{penalty} \gets \textit{prev\_penalty}$
\State $\eta \gets \textit{$\eta_{R_C}$}$
% \State $\textit{tt} \gets \textit{prev\_tt} $
\State $\textit{CT} \gets \textit{prev\_CT} $
\State $\textit{energyRem} \gets \textit{current energy of $R^{r'}_{C}$} $

\If{$(\textit{penalty} > new\_penalty$ or $\textit{penalty} = new\_penalty$ and $\eta > new\_\eta_{C}$ or $\eta = new\_\eta_{C}$ and $\textit{energyRem} > new\_energyRem)$} \label{start least efficient robot}
    \State $\textit{prev\_bid} \gets Bid$ \label{end least efficient robot}
    \State $\textit{prev\_bestrobot} \gets R_{C}^{r'}$
    \State $\textit{prev\_penalty} \gets new\_penalty$
    % \State $\textit{prev\_tt} \gets Bid.tt$
    \State $\textit{prev\_CT} \gets new\_CT$
    \State $\eta \gets new\_\eta_{C}$
    \State $\textit{energyRem} \gets \textit{(current energy of $R^{r'}_{C}$-UsedEnergy of $R^{r'}_{C}$)}$
\EndIf
\State $\textbf{return} \, \textit{BestBid}, \textit{BestRobot}$ 
\end{algorithmic}
\end{algorithm}

\begin{itemize}
        
        \item \textbf{penalty $>$ new\_penalty:} This condition suggests that if the penalty associated with the current bid is lower than that of the best bid, the current bid is deemed the best bid. This approach minimizes potential service request execution delays (refer to Eq. \ref{equation 10}).

        \item \textbf{penalty = new\_penalty and $\eta$ $>$ new\_$\eta_C$:} If the penalty of the current bid matches that of the best bid, and the $\eta$ (a measure of efficiency) for the current bid is lower than that of the best bid, the service request is assigned to the robot with the lower $\eta$. This ensures that even the least efficient robot receives aid to complete service requests within the stipulated time. Opting for the minimum $\eta$ enables the team to assign the most suitable robot to handle service requests that other robots are unable to execute. Therefore, the rejection rate or number of accepted service requests is minimized. 
        
        \item \textbf{$\eta$ = new\_$\eta_C$ and energyRem $>$ new\_energyRem:} This condition suggests that when two bids have the same penalty and efficiency measures but vary in remaining energy levels. Opting for a bid with lower remaining energy is beneficial. This preference stems from the insight that a robot with a higher energy level can be reserved for upcoming service requests with shorter deadlines or hard time constraints. Consequently, this strategy further reduces the overall incurred penalty, as it ensures the availability of a more capable robot for critical service requests.

        \item If the current bid is at least as good as the best bid seen so far, the algorithm updates the following: BestBid becomes the current Bid, update the BestRobot as $R^{r}_{C}$, and $prev\_penalty$, $prev\_CT$, $\eta$, and $energyRem$ are updated with the values from the current Bid. By integrating these considerations into the bidding process, the system can optimize the allocation of service requests among the available robots. This minimizes the total penalty incurred for not completing service requests within the deadline (refer to Eq. \ref{equation 10}). Moreover, the characteristics of the bidding process also minimize the total number of rejected service requests. 
    
    \item Finally, the function returns the best bid (BestBid) and the best robot (BestRobot) found during the comparison.
\end{itemize}

\subsection{Schedule of the Individual Robot (Algorithm \ref{Algorithm5})} 

After determining the winning robot from the auction, the winning robot adds the new service request to its service request list (SRL). Then, it calculates the schedule for all the service requests using the STN (Line \ref{Schedule}). The STN is used to create a schedule that takes into account the time it takes for the robot to reach each node (pickup and drop location). After getting

\begin{algorithm}[htbp]
\caption{IndividualRobotSchedule($ R^{r'}_{C}, SRL_{R^{r'}_{C}}$)} \label{Algorithm5}
\begin{algorithmic}[1]
\State $\textit{nodes} = \{\}$
\State $\textit{currNodes} = \text{set()}$
\State $\textit{currPos} = \text{depot of $ R^{r'}_{C}$}$

\For{$i$ in $0$ to $\text{len}(SRL_{R^{r'}_{C}}) - 1$}
    \State $\textit{nodes}[SRL_{R^{r'}_{C}}[i].j] = SRL_{R^{r'}_{C}}[i]$
    \State $\textit{currNodes}.\text{add}(SRL_{R^{r'}_{C}}[i].j)$
\EndFor

\State $\textit{Schedule} = \text{STN}(\textit{currPos}, \textit{currNodes})$ \label{Schedule}

\While{$\text{size}(\textit{Schedule}) \neq \text{empty}$} \label{start schedule}
    \State $\textit{currPos} = \textit{node}$ \label{update position}
    \State $\textit{node} = \textit{Schedule}.\text{pop}()$ \label{remove node}
    
    \If{$\text{update} \in SRL_{R^{r'}_{C}}$} \label{update in service request list}
        \State $\text{STN}(\textit{currPos}, \textit{currNodes})$
    \EndIf 
\EndWhile \label{end schedule}

\State \textbf{update} (\textit{currPos} $=$ \textit{Schedule[node]}) \label{update}
\end{algorithmic}
\end{algorithm}

\noindent the schedule, the robot moves toward the first node in the schedule and updates its current position (Line \ref{update position}). The robot then removes the previously visited node from the schedule (Line \ref{remove node}). This step ensures that the robot does not revisit nodes it has already serviced. The robot again checks whether it has won any new service requests or not. If there are no new service requests, it continues executing the existing schedule. However, if the robot has won a new service request (Line \ref{update in service request list}), it calculates a new STN schedule that includes the winning service request. This is done to incorporate the new service request into the robot's schedule. Steps \ref{start schedule} to \ref{end schedule} represent a loop where the robot continues to execute its schedule, check for new service requests, and update its schedule accordingly if necessary. At the end (Line \ref{update}), if there are no service requests left in its SRL or it is waiting for a new service request, the robot updates its current position. This likely means that the robot has completed all its assigned service requests or is waiting for new service request(s).

\subsection{Time Complexity}
To understand its efficiency, we analyze the time complexity of each step and determine the overall complexity. The time complexity of auctioneer determination, Algorithm \ref{Algorithm2} is $O(C)$, where $C$ is constant. The bid evaluation process, Algorithm \ref{Algorithm3}, has a time complexity of $O(s^3)$, where $s$ is the number of time nodes in the network. The number of time nodes is at most 2$\times len(SRL)$ for any robot, which is constant due to constraints \ref{equation 6}. Therefore, the time complexity of Algorithm \ref{Algorithm3} is $O(C)$. Selecting the best bid, Algorithm \ref{Algorithm4} involves no loops, so its time complexity is $O(C)$. For scheduling individual robots, Algorithm \ref{Algorithm5} has a time complexity of $O(len(SRL))$, which is also constant due to constraints \ref{equation 6}. Thus, the time complexity of Algorithm \ref{Algorithm5} is $O(C)$. For each task, Algorithm \ref{Algorithm1} iterates over the two nested loops. The first loop (Line \ref{start bidcalculation}) iterates over the total number of robot classes $m$, and the second loop (Line \ref{secondloop}) iterates over the maximum number of robots in a robot class, $n$. At Line \ref{ComputeBid}, Algorithm \ref{Algorithm3} is called, and at Line \ref{BestBid}, Algorithm \ref{Algorithm4} is called. Therefore, the time complexity of Algorithm \ref{Algorithm1} is $O(m)O(n)$, which simplifies to $O(mn)$. The overall time complexity of the system, Algorithm \ref{Algorithm1}, is $O(\lvert J \lvert mn)$+$O(\lvert J \lvert*C)$, simplifying to $O(\lvert J \lvert mn)$, where $\lvert J \lvert$ represents the total number of arrived service requests.

\section{Configuration of Simulation Environment} \label{section 5}

The proposed online allocation algorithm is implemented using the Robot Operating System (ROS) \cite{quigley2009ros} in conjunction with Gazebo \cite{koenig2004design}, making it particularly suitable for dynamic Multi-Robot Task Allocation (MRTA) \cite{dai2020multi} simulations. A controlled simulation environment was designed to assess the algorithm’s performance, leveraging ROS to create a realistic framework. The simulation system employs a distributed programming approach, where each robot runs the complete program independently as separate ROS nodes. This setup ensures that the simulation closely mimics real-world scenarios. Three methods—HMR-ODTA, GA-MR \cite{martin2021multi}, and EEPI \cite{wang2023efficient} were implemented within this system to facilitate comparative analysis. Furthermore, the simulator offers flexibility by allowing customization of parameters such as map size, number of robots, and number of tasks, enabling tailored experiments to evaluate the algorithm’s efficacy.

\subsection{Structure For Configuring Simulation Environment}
    
We utilized the Gazebo simulation environment to model the robots, service requests, and the operational environment. These components were defined and described using the Simulation Description Format (SDF), enabling a detailed and flexible representation within Gazebo. This setup facilitated the evaluation of our allocation algorithm (HMR-ODTA) alongside other recent approaches, such as EEPI \cite{wang2023efficient} and GA-MR \cite{martin2021multi}.

Communication between nodes in this system is achieved through topics and services. The simulation replicates a hospital-like environment where a heterogeneous team of robots collaboratively fulfills service requests. Prior to execution, the simulator initializes several key parameters:
  
\begin{itemize}
    \item The octal distance between map points (potential service request locations) is precomputed using the JPS path planning algorithm. This helps determine distances and aids in efficient path planning for robots.

    \item  The simulated robot features a nonholonomic, four-wheel configuration equipped with a 360-degree LiDAR sensor for environment perception and autonomous navigation.
    
    \item The number of robots in each class is defined to enable the task allocation algorithms to assign suitable robots for specific service requests.
    
    \item The initial positions of the robots are specified.
    
    \item Service requests are dynamically generated at a configurable arrival rate ranging from 0 to 10 seconds.
   
\end{itemize}

\begin{figure}[t]
    \centering
        \includegraphics[width=\linewidth]{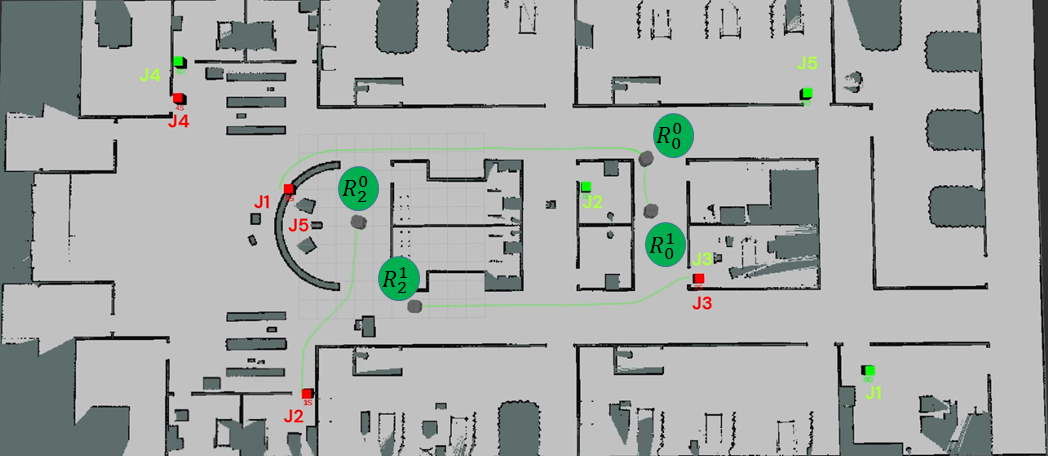}
   
    \caption{The simulation of robot task allocation and navigation within a hospital building involves robots (denoted by \( R_0^0 \), \( R_1^0 \), \( R_1^2 \), etc.) being assigned to complete service requests labeled \( j_1 \), \( j_2 \), \( j_3 \), etc.), at specified locations. Pickup points are marked with red squares, while drop-off points are indicated by green squares. The green paths trace the routes the robots follow as they move between the pickup and drop-off locations. The layout illustrates the hospital floor plan, with robots navigating to complete their designated tasks.} 
    \label{s1}
\end{figure}

All experiments were conducted on an Intel NUC (NUC10i5FNHN) CPU running at 4.20 GHz with 8 GB RAM. Scheduling activities are carried out on a grid world environment measuring $52 \times 25\: m^2$, where each cell spans $1m \times 1m$, as shown in Fig. \ref{Hospital Building}. This simulation system offers a robust platform for testing various task allocation algorithms in a hospital-like scenario, where diverse robots collaboratively handle service requests independently. Fig. \ref{s1} depicts the arrival of several service requests, with robots navigating toward their designated locations. For example, robots $R^0_0$ and $R^0_1$ have been assigned to service requests $j_1$ and $j_5$, respectively.

\subsection{Scenarios} \label{Scenarios}

To evaluate our approach, we conducted simulations across various scenarios using four distinct methods. These scenarios incorporated diverse deadlines for incoming service requests and involved robot classes with either similar or differing attributes. Online service requests were generated dynamically, with their number ranging from 40 to 280, while the total number of robots was fixed at 80. The following deadline configurations (DD) were considered during the simulations:

  \begin{itemize}
    \item $DD = E$ (execution time) [E = (octal distance between pickup and drop off locations)/min velocity]: In this case, the service request deadline is configured equal to its execution time. This can lead to a considerable number of service requests missing their deadline when there are a large number of service requests and relatively few robots available. While this situation is unlikely to occur frequently in a real-world hospital environment setting, it gives us an idea of how the scheduler performs under worst-case conditions. \label{E}
    
    \item $DD = 2E$: In this set, the deadline is set to twice the execution time of the service request. Although this is also a rare occurrence in practical hospital environment scenarios, some service requests may miss their deadline in situations where there are a limited number of robots available. This scenario provides another worst-case view of the scheduler. \label{2E}

    \item $DD \in [$5E$, $10E$]$: In this case, the deadline is set using a uniform distribution in the range of 5E and 10E. This scenario allows for a greater number of service requests to be scheduled within their deadline. \label{5E}
    
    \item $DD \in [E, 10E]$: In this case, the deadline is chosen randomly from a mix of the previous three scenarios with equal probability. This ensures that the scheduler does not favor a particular deadline configuration and provides a more comprehensive assessment of the performance of the scheduler. \label{10E}

\end{itemize}

We adopt the same approach for assigning deadlines and techniques for generating data for real-time scheduling as in \cite{baker2005comparison}. Subsequently, we evaluate the scheduler in two different scenarios for every possible combination of deadlines:  

\begin{itemize} \label{Scenario}

\item \textbf{Scenario 1}: In this scenario, each of the four classes ($CL_0$, $CL_1$,$CL_2$, and $CL_3$) had unique attributes as specified in Table \ref{Table1}, and 20 robots are incorporated in each class. 

 \item \textbf{Scenario 2}:  Different numbers of robots were assigned to each class, but each class had unique attributes shown in Table \ref{Table1}. We ensured that there were 13, 20, 22, and 25 robots in classes $CL_0$, $CL_1$,$CL_2$, and $CL_3$, respectively. 

\end{itemize}

These simulations allowed us to rigorously assess the scheduler’s performance under diverse task deadlines and robot configurations.

\begin{table}[t]
% \centering
\caption{Robot attributes and classes}
\label{Table1} % Fixed label with underscores instead of spaces
\renewcommand{\arraystretch}{1.5} % Adjust the value to add desired spacing
\begin{tabular}{|l|c|c|c|c|}
\hline
\textbf{Robot classes/Attributes} & \boldmath$R_0$ & \boldmath$R_1$ & \boldmath$R_2$ & \boldmath$R_3$ \\ \hline
\textbf{Starting Point} & Dock2 & Dock3 & Dock2 & Dock1 \\ \hline
\textbf{v (m/s)}  & 1.5 & 1 & 0.75 & 0.5 \\ \hline
\textbf{C (kg)}  & 60 & 75 & 85 & 95 \\ \hline
\textbf{E (J)}  & 4000 & 5000 & 6500 & 7000 \\ \hline
\textbf{W (kg)}  & 90 & 80 & 70 & 60 \\ \hline
\end{tabular}
\end{table}

\section{Experiment and Results} \label{section 6}

\begin{table}[t]
\caption{Mapping of Robot Classes and Types of Service Requests}
  \label{Table2}
  \renewcommand{\arraystretch}{1.5} % Adjust the value to add desired spacing
  \centering
  
  \begin{tabular}{|p{1.5cm}|p{9cm}|}
    \hline
    \textbf{Robot Class} & \textbf{Types of service request that can be assigned to $m^{th}$ class robot} \\
    \hline
    $CL_{0}$ & \{$T_{1}$, $T_{2}$, $T_{3}$, $_{4}$, $T_{y5}$, $T_{y6}$, and $T_{y7}$\} \\
    \hline
    $CL_{1}$ & \{$T_{1}$, $T_{3}$, and $T_{4}$\} \\
    \hline
    $CL_{2}$ & \{$T_{1}$, and $T_{4}$\} \\
    \hline
    $CL_{3}$ & \{$T_{6}$, and $T_{7}$\} \\
    \hline
  \end{tabular}
\end{table}

The proposed approach for online task allocation in heterogeneous multi-robot systems, HMR-ODTA, was extensively evaluated through comprehensive simulation experiments conducted in a hospital building environment (refer to Fig.  \ref{Hospital Building}). For robust comparison, two state-of-the-art (SOTA) approaches, GA-MR \cite{martin2021multi} and EEPI \cite{wang2023efficient}, were re-implemented and rigorously tested under identical conditions. These experiments, focused on the online pickup-and-delivery problem, involved 50 simulation runs to ensure statistical significance, minimizing the influence of chance or random variations on the findings. The GA-MR algorithm was adapted to the online pickup-and-delivery problem by incorporating allocation techniques from the original algorithm, where service request allocation considered both estimated completion time and remaining fuel levels. Similarly, the EEPI algorithm was implemented with heterogeneous robots, assuming equal priority for all service requests, resulting in uniform penalties for delay. EEPI emphasized scheduling requests to maximize successfully accepted tasks, with task allocation based on the robot’s start time and execution time to meet task deadlines. These evaluations validate the proposed algorithm’s robustness and effectiveness in handling real-time challenges compared to GA-MR and EEPI under identical conditions.

\subsection{Results and Discussions}

We conducted 50 simulation runs for both the two re-implemented SOTA approaches and our proposed HMR-ODTA approach. The simulations varied the number of service requests $(\lvert J \lvert)$ from 40 to 280, with 50 runs performed for each value of $(\lvert J \lvert)$. These service requests, submitted by users, were assigned limited delivery time windows and were generated as described in Section \ref{Scenarios}. The time constraints were intentionally stringent, with some deadlines being physically impractical to meet, emphasizing our primary goal of optimizing the delivery process as much as possible. Four robot classes were considered in the simulations (refer to Table \ref{Table2}), where each class could handle specific types of service requests (e.g., robot class $CL_2$ could perform tasks of types $T_{1}$ and $T_{4}$, as listed in Table \ref{Table3}). To ensure fairness, the starting positions of robots were kept consistent across all approaches. Simulation parameters were configured as described in \cite{Huaicheng}. We evaluated each approach using two metrics: (a) Cumulative Incurred Penalty, measuring the time taken by the robot team to complete accepted service requests beyond their deadlines, and (b) Cumulative Rejected Service Requests, representing the total number of service requests not accepted by the robots.

\begin{table}[t]
\caption{Service request types and corresponding required abilities}
  \label{Table3}
  \renewcommand{\arraystretch}{1.5} % Adjust the value to add desired spacing
  \centering
  
  \begin{tabular}{|p{2cm}|p{7cm}|}
    \hline
    \textbf{Service request type} & \textbf{Description} \\
    \hline
    $T_{1}$ & Delivery of Supplies\\
    \hline
    $T_{2} $ &  Telemedicine\\
    \hline
    $T_{3} $ &  Monitoring and Data Collection\\
    \hline
    $T_{4} $ & Patient Assistance \\
    \hline
    $T_{5} $ & Autonomous Transportation \\
    \hline
    $T_{6} $ & Logistics and Inventory Management \\
    \hline
    $T_{7} $ & Disinfection and Cleaning \\
    \hline
  \end{tabular}  
\end{table}

\subsection{The Algorithm's Performance Under Different Scenarios}

\begin{figure*}[htbp]
    \centering
    \begin{subfigure}[t]{0.49\linewidth}
        \centering
        \includegraphics[width=0.9\linewidth]{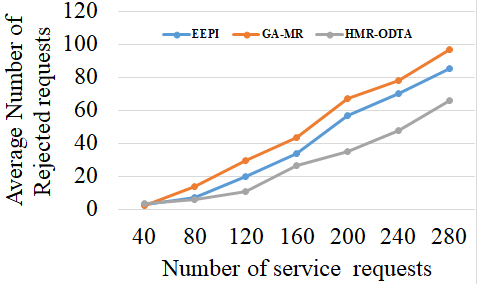}
        \caption{Equal robot (ER) with unique attributes (UA) with the deadline (E)} 
        \label{ERUA-E}
    \end{subfigure}
    \hfill
    \begin{subfigure}[t]{0.49\linewidth}
        \centering
        \includegraphics[width=0.9\linewidth]{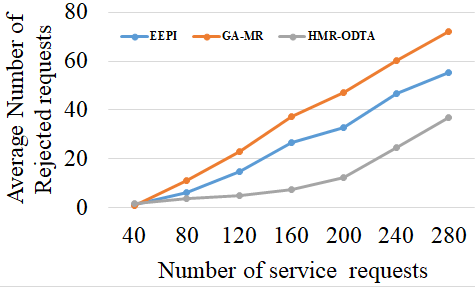}
        \caption{Equal robot (ER) with unique attributes (UA) with the deadline (2E)} 
        \label{ERUA-2E}
    \end{subfigure}
    \begin{subfigure}[t]{0.49\linewidth}
        \centering
        \includegraphics[width=0.9\linewidth]{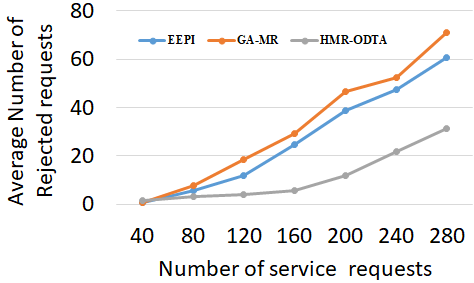}
        \caption{Equal robot (ER) with unique attributes (UA) with the deadline (5E, 10E)} 
        \label{ERUA-5E}
    \end{subfigure}
    \hfill
    \begin{subfigure}[t]{0.49\linewidth}
        \centering
        \includegraphics[width=0.9\linewidth]{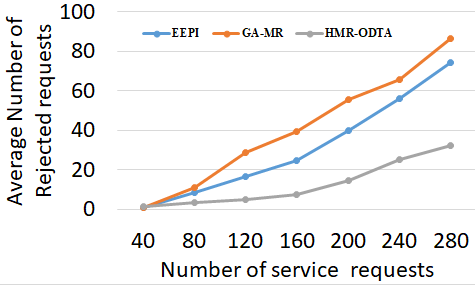}
        \caption{Equal robot (ER) with unique attributes (UA) with the deadline (E, 10E)} 
        \label{ERUA-10E}
    \end{subfigure}
    \begin{subfigure}[t]{0.49\linewidth}
        \centering
        \includegraphics[width=0.9\linewidth]{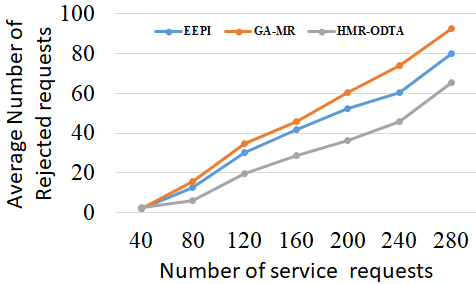}
        \caption{Unequal robot (UR) and unique attributes (UA) with the deadline (E)}
        \label{URUA-E}
    \end{subfigure}
    \hfill
    \begin{subfigure}[t]{0.49\linewidth}
        \centering
        \includegraphics[width=0.9\linewidth]{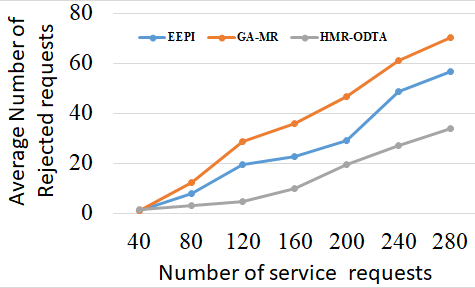}
        \caption{Unequal robot (UR) and unique attributes (UA) with the deadline (2E)}
        \label{URUA-2E}
    \end{subfigure}
    \begin{subfigure}[t]{0.49\linewidth}
        \centering
        \includegraphics[width=0.9\linewidth]{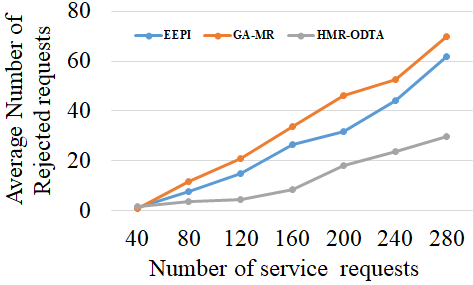}
        \caption{Unequal robot (UR) and unique attributes (UA) with the deadline (5E, 10E)}
        \label{URUA-5E}
    \end{subfigure}
    \hfill
    \begin{subfigure}[t]{0.49\linewidth}
        \centering
        \includegraphics[width=0.9\linewidth]{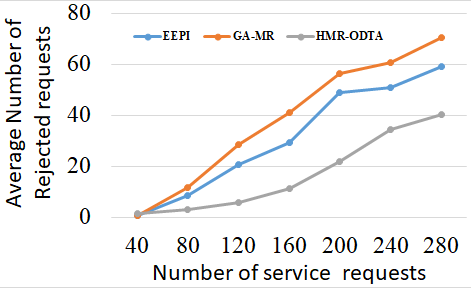}
        \caption{Unequal robot (UR) and unique attributes (UA) with the deadline (E, 10E)}
        \label{URUA-10E}
    \end{subfigure}
    \caption{Number of Service Requests vs. Average Number of Rejected Requests. Robot Team Size = 80. Service requests are varying from 40 to 280 in increments of 40.} 
    
    \label{Rejected-Requests}
\end{figure*}

\begin{figure*}[htbp]
    \centering
    \begin{subfigure}[t]{0.49\linewidth}
        \centering
        \includegraphics[width=0.9\linewidth]{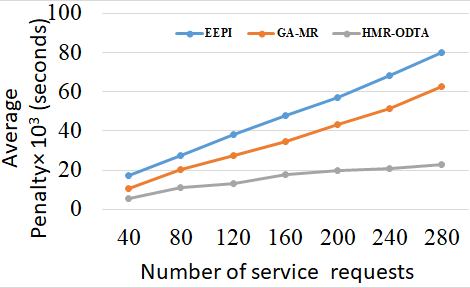}
        \caption{Equal robot (ER) with unique attributes (UA) with the deadline (E)}
        \label{P-ERUA-E}
    \end{subfigure}
    \hfill
    \begin{subfigure}[t]{0.49\linewidth}
        \centering
        \includegraphics[width=0.9\linewidth]{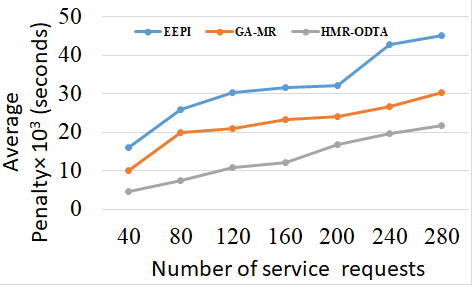}
        \caption{Equal robot (ER) with unique attributes (UA) with the deadline (2E)}
        \label{P-ERUA-2E}
    \end{subfigure}
    \begin{subfigure}[t]{0.49\linewidth}
        \centering
        \includegraphics[width=0.9\linewidth]{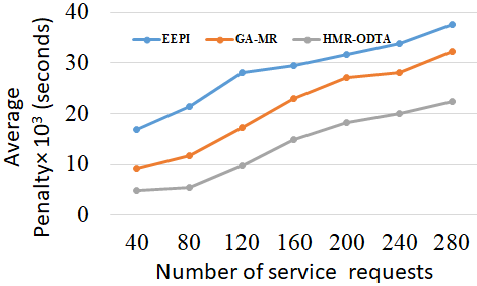}
        \caption{Equal robot (ER) with unique attributes (UA) with the deadline (5E, 10E)}
        \label{P-ERUA-5E}
    \end{subfigure}
    \hfill
    \begin{subfigure}[t]{0.49\linewidth}
        \centering
        \includegraphics[width=0.9\linewidth]{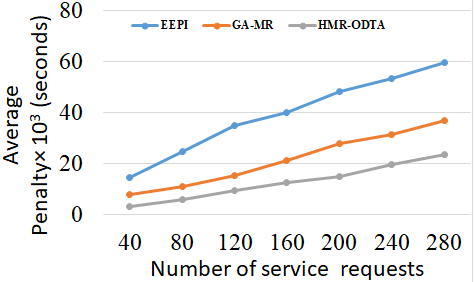}
        \caption{Equal robot (ER) with unique attributes (UA) with the deadline (E, 10E)}
        \label{P-ERUA-10E}
    \end{subfigure}
    \begin{subfigure}[t]{0.49\linewidth}
        \centering
        \includegraphics[width=0.9\linewidth]{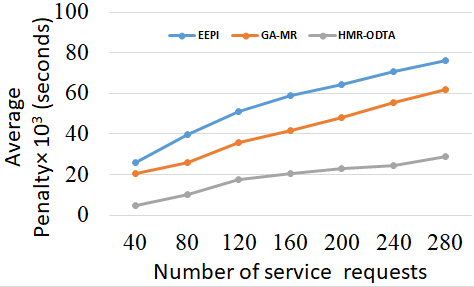}
        \caption{Unequal robot (UR) and unique attributes (UA) with the deadline (E)}
        \label{P-URUA-E}
    \end{subfigure}
    \hfill
    \begin{subfigure}[t]{0.49\linewidth}
        \centering
        \includegraphics[width=0.9\linewidth]{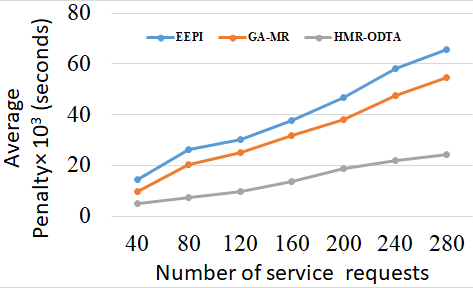}
        \caption{Unequal robot (UR) and unique attributes (UA) with the deadline (2E)}
        \label{P-URUA-2E}
    \end{subfigure}
    \begin{subfigure}[t]{0.49\linewidth}
        \centering
        \includegraphics[width=0.9\linewidth]{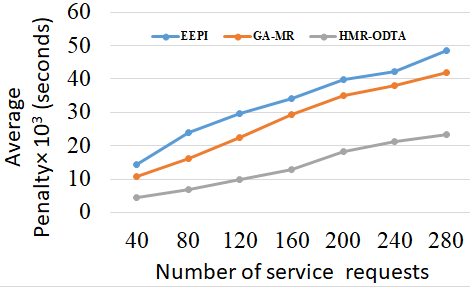}
        \caption{Unequal robot (UR) and unique attributes (UA) with the deadline (5E, 10E)}
        \label{P-URUA-5E}
    \end{subfigure}
    \hfill
    \begin{subfigure}[t]{0.49\linewidth}
        \centering
        \includegraphics[width=0.9\linewidth]{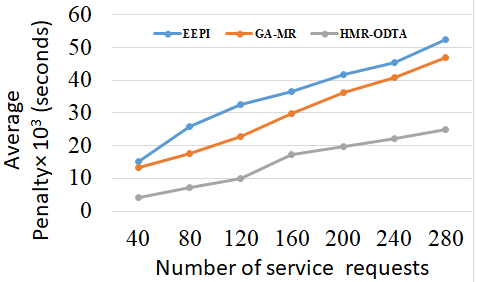}
        \caption{Unequal robot (UR) and unique attributes (UA) with the deadline (E, 10E)}
        \label{P-URUA-10E}
    \end{subfigure}
    
    \caption{ Number of Service Requests vs. Average Penalty (in seconds). Robot Team Size = 80. Service requests are varying from 40 to 280 in increments of 40.} 
    
    \label{Penalty}
\end{figure*}

We fixed the total number of robots at \(80\) and generated schedules for \(50\) trials in each scenario, as shown in Table \ref{Table4}. For each scenario, we evaluated the algorithm's performance based on the number of service requests (\( \lvert J \rvert \)). The service requests were dynamically generated, with a new request appearing every \(0\) to \(10\) seconds. The deadlines for each service request were either uniform or varied and were determined using the method outlined in \cite{baker2005comparison}. This approach ensures that an available robot can pick up the service request from its designated pickup location and deliver it to the drop-off location within the specified deadline. In some cases, robots were unable to complete a service request within the given time window, leading to penalties (for soft time constraints) or rejected requests (for hard time constraints). To handle incoming service requests in real-time, we accounted for the heterogeneity of the robot team and assigned new requests to them accordingly. Initially, the number of service requests are lower than the number of available robots, resulting in a lower average number of rejected requests and penalties. The average number of rejected requests and the total penalty time (in seconds) for each algorithm in delivering the dynamically generated service requests are presented in Fig. \ref{Rejected-Requests} and Fig. \ref{Penalty}, respectively. From Fig. \ref{Rejected-Requests}~(\subref{ERUA-E}) to Fig. \ref{Rejected-Requests}~(\subref{URUA-10E}), it is evident that the EEPI algorithm performs better than GA-MR, while HMR-ODTA delivers the best overall performance. The relatively poor performance of GA-MR is due to the fact that, according to the algorithm, each robot only begins moving towards the next service request after completing the current request and delivering it to its destination. This approach does not fully optimize the robot's capacity and energy. 
%%%

\begin{table}[t]
\centering
\caption{The total number of experiments conducted for the three algorithms—EEPI, GA-MR, and HMR-ODTA—per scenario were evaluated across service request variations ranging from 40 to 280, in increments of 40}
\begin{tabular}{|p{1cm}|p{0.8cm}|p{0.8cm}|p{0.8cm}|p{0.8cm}|p{1.5cm}|p{1.7cm}|p{1.8cm}|}
\hline
\begin{tabular}[c]{@{}l@{}}Scenario\end{tabular} & \begin{tabular}[c]{@{}l@{}}DD=\\ E\end{tabular} & \begin{tabular}[c]{@{}l@{}}DD=\\ 2E\end{tabular} & \begin{tabular}[c]{@{}l@{}}DD=\\ (5E,\\ 10E)\end{tabular} & \begin{tabular}[c]{@{}l@{}}DD=\\ (E,\\ 10E)\end{tabular} & \begin{tabular}[c]{@{}l@{}}Variability\\ of Service \\ Requests \end{tabular} & \begin{tabular}[c]{@{}l@{}}Algorithm\\ count\end{tabular} & \begin{tabular}[c]{@{}l@{}}Total\\ Experiments\end{tabular} \\ \hline
\begin{tabular}[c]{@{}l@{}}ER-UA\end{tabular} & 50 & 50 & 50 & 50 & 7 & 3 & 4200 \\ \hline
\begin{tabular}[c]{@{}l@{}}UR-UA\end{tabular} & 50 & 50 & 50 & 50 & 7 & 3 & 4200 \\ \hline
\end{tabular}
\label{Table4}
\end{table}

In contrast, EEPI uses a proposed cost function that assigns tasks to robots considering multiple criteria, with a primary focus on aligning a robot's start time with the service request's deadline. This strategy aims to ensure that tasks are assigned to robots that can start close enough to the deadlines, reducing the likelihood of missing them due to delayed starts. However, in EEPI, once a task is assigned to a robot based on the converged state, the allocation remains fixed. Unlike dynamic scheduling methods, EEPI’s cost function does not allow for adjustments after the task assignment phase. This becomes problematic when subsequent tasks have tighter deadlines or when the initial assignments are sub-optimal due to unforeseen changes. If a task cannot be completed by the initially assigned robot due to scheduling conflicts or mismatched abilities, the task is rejected, even if other robots could potentially complete it on time. As a result, robot resources can be underutilized, and task completion efficiency may decrease, especially under dynamic conditions. The inability of EEPI’s cost function to reschedule tasks highlights the need for adaptive scheduling algorithms that can dynamically reallocate tasks based on real-time changes in task priorities, robot availability, and operational constraints. 

In contrast, HMR-ODTA adopts a more adaptive approach by using a Simple Temporal Network (STN) framework to assess the availability of robots within the specified time interval $[j_{AT}, j_{DD}]$. This method not only addresses temporal constraints but also integrates critical factors to enhance system efficiency. The framework allows for rescheduling tasks within a robot’s schedule. If a robot experiences a delay or encounters unforeseen tasks, it can automatically reschedule its accepted tasks to minimize disruption and ensure the timely completion of service requests. This adaptive rescheduling capability is crucial for maintaining high service reliability. Additionally, the framework considers the types and number of service requests that can be assigned to each robot. By continuously monitoring and analyzing robot performance, HMR-ODTA can dynamically adjust task assignments, ensuring optimal utilization of the heterogeneous robot team. The primary goal of HMR-ODTA is to maximize the number of completed service requests while minimizing associated penalties. By integrating these factors into the STN framework, the system effectively balances trade-offs between energy consumption, the range of service requests each robot can handle, and the need for rescheduling. 

\subsection{Rejected Requests Analysis}
The Fig. \ref{Rejected-Requests}~(\subref{ERUA-E}) - Fig. \ref{Rejected-Requests}~(\subref{URUA-10E}), presents a comparison of three different task scheduling algorithms EEPI, GA-MR, and HMR-ODTA based on their performance across varying numbers of requests. The average number of rejected requests serves as an indicator of the system’s ability to accommodate the demand. A higher number of rejected requests suggests the algorithm is less capable of handling the given load, leading to potential service failures or unmet demands.
\begin{table}[t]
\centering
\caption{Percentage difference of number of rejected service requests between the EEPI, GA-MR, and HMR-ODTA for the service requests ranging from 40 to 160 and 160 to 280.}
\begin{tabular}{|c|l|ll|ll|}
\hline
\multirow{2}{*}{Deadline} &
  \multicolumn{1}{c|}{\multirow{2}{*}{\begin{tabular}[c]{@{}c@{}}\% difference between \\ algorithms\end{tabular}}} &
  \multicolumn{2}{c|}{UR-UA} &
  \multicolumn{2}{c|}{ER-UA} \\ 
                        & \multicolumn{1}{c|}{} & \multicolumn{1}{l|}{40-160} & 160-280 & \multicolumn{1}{l|}{40-160} & 160-280 \\ \hline
\multirow{3}{*}{E}      & EEPI vs GA-MR         & \multicolumn{1}{l|}{12.14} & 15.91   & \multicolumn{1}{l|}{34.34} & 13.22   \\  
                        & EEPI vs HMR-ODTA      & \multicolumn{1}{l|}{51.81} & 42.21   & \multicolumn{1}{l|}{63.32} & 48.09   \\  
                        & GA-MR vs HMR-ODTA     & \multicolumn{1}{l|}{40.30} & 26.74   & \multicolumn{1}{l|}{30.64} & 35.44   \\ \hline
\multirow{3}{*}{2E}     & EEPI vs GA-MR         & \multicolumn{1}{l|}{40.98} & 27.84   & \multicolumn{1}{l|}{38.66} & 28.10   \\  
                        & EEPI vs HMR-ODTA      & \multicolumn{1}{l|}{120.15} & 75.61  & \multicolumn{1}{l|}{122.35} & 83.77  \\  
                        & GA-MR vs HMR-ODTA     & \multicolumn{1}{l|}{90.28} & 50.42   & \multicolumn{1}{l|}{94.91} & 59.15   \\ \hline
\multirow{3}{*}{5E,10E} & EEPI vs GA-MR         & \multicolumn{1}{l|}{29.91} & 19.86   & \multicolumn{1}{l|}{26.33} & 14.26   \\  
                        & EEPI vs HMR-ODTA      & \multicolumn{1}{l|}{114.90} & 81.22  & \multicolumn{1}{l|}{118.89} & 89.31  \\  
                        & GA-MR vs HMR-ODTA     & \multicolumn{1}{l|}{92.98} & 63.94   & \multicolumn{1}{l|}{100.42} & 77.51  \\ \hline
\multirow{3}{*}{E,10E}  & EEPI vs GA-MR         & \multicolumn{1}{l|}{31.98} & 16.74   & \multicolumn{1}{l|}{44.81} & 19.71   \\  
                        & EEPI vs HMR-ODTA      & \multicolumn{1}{l|}{116.20} & 64.15  & \multicolumn{1}{l|}{127.76} & 97.23  \\  
                        & GA-MR vs HMR-ODTA     & \multicolumn{1}{l|}{92.85} & 48.72   & \multicolumn{1}{l|}{96.80} & 81.42   \\ \hline
\end{tabular}
\label{Table5}
\end{table}

\begin{itemize}
    \item EEPI and GA-MR: Both algorithms exhibit a significant increase in rejected requests as the number of requests grows. GA-MR, in particular, shows a sharp rise, especially at higher request levels (240 and 280) (refer to Fig. \ref{Rejected-Requests} ), indicating it's declining performance under increased load. EEPI, while slightly better than GA-MR, also shows a substantial number of rejections, highlighting its limitations in managing higher request volumes. These limitations are likely due to their less flexible task allocation strategies, which cannot effectively cope with the increasing complexity and volume of tasks. EEPI outperforms GA-MR by nearly 12.14\% for a smaller number of service requests, ranging from 40 to 160. For larger instances, with service requests ranging from 160 to 280, GA-MR shows an improvement of 15.91\% over GA-MR, as shown in Table \ref{Table5}. This performance difference is observed under very tight deadlines, i.e., E.
    
    \item HMR-ODTA: This algorithm demonstrates a markedly lower number of rejected requests at all levels. Even as the number of requests increases, the growth in rejected requests is more controlled and gradual, as can be seen in Fig. \ref{Rejected-Requests}~\subref{ERUA-E} - Fig. \ref{Rejected-Requests}~\subref{URUA-10E}. This indicates HMR-ODTA’s robust capability in handling a larger number of requests effectively, minimizing service rejections. The reason for this robustness is its ability to dynamically reschedule tasks and optimize resource utilization in real-time, accommodating more requests without significant degradation in performance. HMR-ODTA outperforms EEPI and GA-MR by nearly 51.81\% and 40.30\%, respectively, for a smaller number of service requests, ranging from 40 to 160. For larger instances, with service requests ranging from 160 to 280, HMR-ODTA shows an improvement of 42.21\% and 26.74\% over EEPI and GA-MR, respectively, as shown in Table \ref{Table5}.
\end{itemize}

\subsection{Penalty Analysis}

The penalty metric reflects the system's efficiency in handling requests within the given constraints. Lower penalty values indicate better performance in terms of timely completion of tasks. It can be seen from Fig. \ref{Penalty}~(\subref{P-ERUA-E}) - Fig. \ref{Penalty}~(\subref{P-URUA-10E}).

\begin{itemize}
    \item EEPI and GA-MR: These algorithms exhibit higher penalty values across all request levels, indicating less efficient handling of task scheduling compared to HMR-ODTA. As the number of requests increases, the penalty for EEPI and GA-MR also increases, showing a linear to slightly exponential trend, which suggests these algorithms struggle to maintain efficiency under heavier loads. This can be attributed to their less adaptive scheduling mechanisms that are not as capable of dynamically responding to changes in the environment or workload. EEPI and GA-MR rely on static allocation strategies that do not consider real-time data, leading to suboptimal performance as task complexity and volume grow. GA-MR outperforms EEPI by nearly 33.19\% for a smaller number of service requests, ranging from 40 to 160. For larger instances, with service requests ranging from 160 to 280, GA-MR shows an improvement of 38.89\% over EEPI,as shown in Table \ref{Table6}. This performance difference is observed under very tight deadlines, i.e., E. Although the \% difference of GA-MR is better than the EEPI, the main reason behind it is the number of rejected requests in GA-MR is more. If more requests are rejected, that will lead to a reduction of the penalty.

\begin{table}[t]
\centering
\caption{Percentage difference of the reduction of cumulative penalties of service requests between the EEPI, GA-MR, and HMR-ODTA for the service requests ranging from 40 to 160 and 160 to 280.}
\begin{tabular}{|c|l|ll|ll|}
\hline
\multirow{2}{*}{Deadline} &
  \multicolumn{1}{c|}{\multirow{2}{*}{\begin{tabular}[c]{@{}c@{}}\% difference between \\ algorithms\end{tabular}}} &
  \multicolumn{2}{c|}{UR-UA} &
  \multicolumn{2}{c|}{ER-UA} \\ 
                        & \multicolumn{1}{c|}{} & \multicolumn{1}{l|}{40-160} & \multicolumn{1}{l|}{160-280} & \multicolumn{1}{l|}{40-160} & \multicolumn{1}{l|}{160-280} \\ \hline
\multirow{3}{*}{E}      & EEPI vs GA-MR         & \multicolumn{1}{l|}{33.19}  & 38.89   & \multicolumn{1}{l|}{34.43}  & 24.46   \\  
                        & EEPI vs HMR-ODTA      & \multicolumn{1}{l|}{71.74}  & 32.94   & \multicolumn{1}{l|}{80.47}  & 73.23   \\  
                        & GA-MR vs HMR-ODTA     & \multicolumn{1}{l|}{99.03}  & 69.60   & \multicolumn{1}{l|}{107.45} & 93.50   \\ \hline
\multirow{3}{*}{2E}     & EEPI vs GA-MR         & \multicolumn{1}{l|}{33.88}  & 26.29   & \multicolumn{1}{l|}{22.30}  & 19.29   \\  
                        & EEPI vs HMR-ODTA      & \multicolumn{1}{l|}{63.89}  & 85.04   & \multicolumn{1}{l|}{84.37}  & 73.30   \\  
                        & GA-MR vs HMR-ODTA     & \multicolumn{1}{l|}{92.75}  & 105.43  & \multicolumn{1}{l|}{101.88} & 89.43   \\ \hline
\multirow{3}{*}{5E,10E} & EEPI vs GA-MR         & \multicolumn{1}{l|}{44.36}  & 16.30   & \multicolumn{1}{l|}{25.87}  & 13.19   \\  
                        & EEPI vs HMR-ODTA      & \multicolumn{1}{l|}{55.10}  & 36.37   & \multicolumn{1}{l|}{79.33}  & 58.58   \\  
                        & GA-MR vs HMR-ODTA     & \multicolumn{1}{l|}{93.73}  & 51.91   & \multicolumn{1}{l|}{100.07} & 70.41   \\ \hline
\multirow{3}{*}{E,10E}  & EEPI vs GA-MR         & \multicolumn{1}{l|}{69.34}  & 50.44   & \multicolumn{1}{l|}{27.61}  & 11.79   \\  
                        & EEPI vs HMR-ODTA      & \multicolumn{1}{l|}{55.57}  & 49.79   & \multicolumn{1}{l|}{73.17}  & 59.80   \\  
                        & GA-MR vs HMR-ODTA     & \multicolumn{1}{l|}{113.93} & 94.31   & \multicolumn{1}{l|}{95.93}  & 70.35   \\ \hline
\end{tabular}
\label{Table6}
\end{table}

    \item HMR-ODTA: This algorithm consistently shows the lowest penalty values across all request levels. The increase in penalties with the number of requests is more gradual than that of EEPI and GA-MR, demonstrating HMR-ODTA’s superior capability in managing larger volumes of tasks efficiently. This efficiency stems from its adaptive approach, which optimizes task allocation based on real-time data and constraints, minimizing delays. HMR-ODTA likely employs dynamic scheduling techniques that continuously adjust to current conditions, resulting in more effective task management. HMR-ODTA outperforms EEPI and GA-MR by nearly 71.74\% and 99.03\%, respectively, for a smaller number of service requests, ranging from 40 to 160. For larger instances, with service requests ranging from 160 to 280, HMR-ODTA shows an improvement of 32.94\% and 69.60\% over EEPI and GA-MR, respectively, as shown in Table \ref{Table6}.
This performance difference is observed under very tight deadlines, i.e.,
E.

    The reasons for this superior performance include:
    \begin{itemize}
        \item Real-Time Data Utilization: HMR-ODTA leverages real-time data to make informed decisions about task allocation and scheduling. This allows the system to adjust quickly to changing conditions and demands.
        \item Dynamic Scheduling: The algorithm uses dynamic scheduling methods that can adapt to variations in task requirements and robot availability. This flexibility helps in maintaining low penalties even as the number of tasks increases.
        \item Resource Utilization: HMR-ODTA optimizes the utilization of available resources, ensuring that each robot is used to its fullest potential without overloading any single unit. This balanced approach reduces the delays and penalties.
    \end{itemize}
\end{itemize}

\subsection {Significance Level Analysis Across Different Scenarios}

In our study, we conducted a rigorous comparison of algorithm performance using the two-tailed Wilcoxon signed-rank test at a significance level of \(5\%\). This test is particularly suited for non-parametric analysis, allowing us to assess whether the distributions of two independent groups significantly differ.

We specifically analyzed small instances where $\lvert J \lvert$ varied from 40 to 160. In one notable case, the effectiveness of HMR-ODTA did not exhibit significance, which we detail as follows:
\begin{itemize}
    \item Average penalty for unequal robots (UR) with unique attributes (UA) where \(DD \in [2E]\).
\end{itemize}

In summary, our results consistently show that HMR-ODTA surpasses Multi-robot task allocation using Genetic Algorithm (GA-MR) and Effective and Efficient Performance Impact (EEPI) in all tested scenarios. This superiority is especially notable in scenarios with high volumes of service requests, highlighting the robustness and effectiveness of HMR-ODTA in addressing real-world optimization challenges.

\section{Conclusion} \label{section 7}

This paper presents HMR-ODTA, a new decentralized auction-based algorithm for task assignment specifically designed for scheduling pickup and delivery tasks with transfers. The algorithm handles online task arrivals, time windows, and task heterogeneity, enabling dynamic coordination of a diverse robot team and real-time schedule adjustments as new tasks emerge. Experimental results in large-scale simulated environments show that HMR-ODTA significantly outperforms state-of-the-art (SOTA) methods. For smaller task sets (40-160 tasks), HMR-ODTA reduces overall penalties by about 63\%, surpassing other approaches. As the task set size grows (160-280 tasks), it further decreases penalties by around 50\%, highlighting its scalability and robustness. This performance is driven by HMR-ODTA's integration of an adaptive task rescheduling mechanism into a robot's schedule, enabling it to manage delays or unknown tasks with minimal disruption while ensuring the timely completion of service requests. Moreover, unlike SOTA methods that typically emphasize a single metric, HMR-ODTA incorporates multi-criteria decision-making, improving team efficiency across different task complexities and robot capabilities. Future work will focus on implementing HMR-ODTA with a real-world team of mobile robots, addressing challenges such as robot and communication failures to validate its effectiveness in practical applications.

\bibliography{HMR-ODTA}

\end{document}